\documentclass[sigconf]{acmart}

\usepackage{float}
\usepackage{mathtools}
\usepackage{epsfig}
\usepackage{verbatim}

\usepackage{amssymb}
\usepackage{amsfonts}
\usepackage{dsfont}
\usepackage{amsbsy}
\usepackage{xcolor}

\usepackage{bbm}
\usepackage{microtype}
\usepackage{graphicx}
\usepackage{subfigure}
\usepackage{booktabs}
\usepackage{hyperref}

\usepackage{algorithm}
\usepackage{algcompatible}

\usepackage{amsmath}
\AtBeginDocument{%
  \providecommand\BibTeX{{%
    \normalfont B\kern-0.5em{\scshape i\kern-0.25em b}\kern-0.8em\TeX}}}

\begin{document}

\title{Wasserstein Graph Neural Networks for\\ Graphs with Missing Attributes}

\author{Zhixian Chen}
\email{zchencz@connect.ust.hk}
\affiliation{%
  \institution{Hong Kong University of Science and Technology}
  \state{Hong Kong SAR}
  \country{China}
}

\author{Tengfei Ma}
\email{Tengfei.Ma1@ibm.com}
\affiliation{%
  \institution{IBM T. J. Watson Research Center}
  \city{New York}
  \country{USA}}

\author{Yangqiu Song}
\author{Yang Wang}
\email{yqsong@cse.ust.hk, yangwang@ust.hk}
\affiliation{%
  \institution{Hong Kong University of Science and Technology}
  \state{Hong Kong SAR}
  \country{China}
}


\begin{abstract}
  Missing node attributes is a common problem in real-world graphs. Graph neural networks have been demonstrated power in graph representation learning while their performance is affected by the completeness of graph information. Most of them are not specified for missing-attribute graphs and fail to leverage incomplete attribute information effectively. In this paper, we propose an innovative node representation learning framework, Wasserstein Graph Neural Network (\textrm{WGNN}), to mitigate the problem. To make the most of limited observed attribute information and capture the uncertainty caused by missing values, we express nodes as low-dimensional distributions derived from the decomposition of the attribute matrix. Furthermore, we strengthen the expressiveness of representations by developing a novel message passing schema that aggregates distributional information from neighbors in the Wasserstein space. We test \textrm{WGNN} in node classification tasks under two missing-attribute cases on both synthetic and real-world datasets. In addition, we find \textrm{WGNN} suitable to recover missing values and adapt them to tackle matrix completion problems with graphs of users and items. Experimental results on both tasks demonstrate the superiority of our method.
\end{abstract}




\maketitle

\section{Introduction}
\label{sec:introduction}

Graphs are ubiquitous data structures, where nodes usually have associated attributes. There have been many impressive and practical machine learning methods on graphs, particularly, graph representation learning \cite{perozzi2014deepwalk,grover2016node2vec,gilmer2017neural} which attempts to embed local structural and attribute information into node representations. Graph representation learning methods underlay various downstream graph-based learning tasks such as node classification and link prediction while the accuracy of most applications is affected by the completeness of graph data. Unfortunately, missing attributes are common in many real-world graph data. For instance, in social networks such as Facebook and Twitter, users tend to hide or selectively publish their personal information for privacy concerns. 
In molecular networks like protein-protein association networks \cite{szklarczyk2019string}, where nodes represent proteins and edges indicate biological associations, it is difficult to obtain complete information about the sequences and molecular structures of newly discovered proteins.
In general, we can categorize node attribute missing issues into two cases: 1).
\textit{Entirely missing}: missing entire attributes on some nodes, 2). \textit{Partially missing}: missing partial attributes on all nodes. In this paper, we investigate graph learning on graphs with missing-attribute. 

 Existing graph representation learning methods are not specified for missing-attribute graphs. Random-walk based network embedding approaches \cite{perozzi2014deepwalk,grover2016node2vec} exploit graph structure information to preserve pre-specified node similarities in the embedding space without considering informative node attributes. Message-passing \cite{gilmer2017neural} based graph neural networks (GNNs) incorporate node attributes and graph structure effectively by aggregating information from neighborhoods. However, to handle incomplete node attributes, they usually need to leverage matrix imputation techniques \cite{troyanskaya2001missing,hastie2015matrix} for missing values estimation before learning. Despite missing value imputation (\textrm{MVI}) being a well-studied problem in data mining and analysis, it is still a big challenge to recover missing values with inadequate observed information \cite{dong2013principled,lin2020missing}. Moreover, imputation methods might introduce noise in representations and restrict graph embedding approaches to model uncertainty caused by missing attributes.

 In this paper, we propose a significant assumption on data: \textit{attributes of each node are sampled from low-dimensional mixture distributions}, similar to the low-rank assumption in most matrix imputation algorithms. More precisely, we assume that there are some latent factors of nodes and each node has a distribution over these latent factors, called node-factor distribution. Meanwhile, each latent factor has a distribution over node attributes, called factor-attribute distribution. For instance, in text mining where nodes represent documents with words as their attributes, topics are their latent factors \cite{arora2012learning}. In practice, the number of (principal) latent factors is usually small. Following this assumption, we employ a matrix-factorization-based method to obtain latent factors from observed node attributes. Thereby, we can also get the node-factor distributions of each node which are exactly low-dimensional node distributional representations. In this way, we can capture the uncertainty caused by incomplete attribute information.
 
\begin{figure*}[!ht]
 \centering
\includegraphics[width=0.87\textwidth]{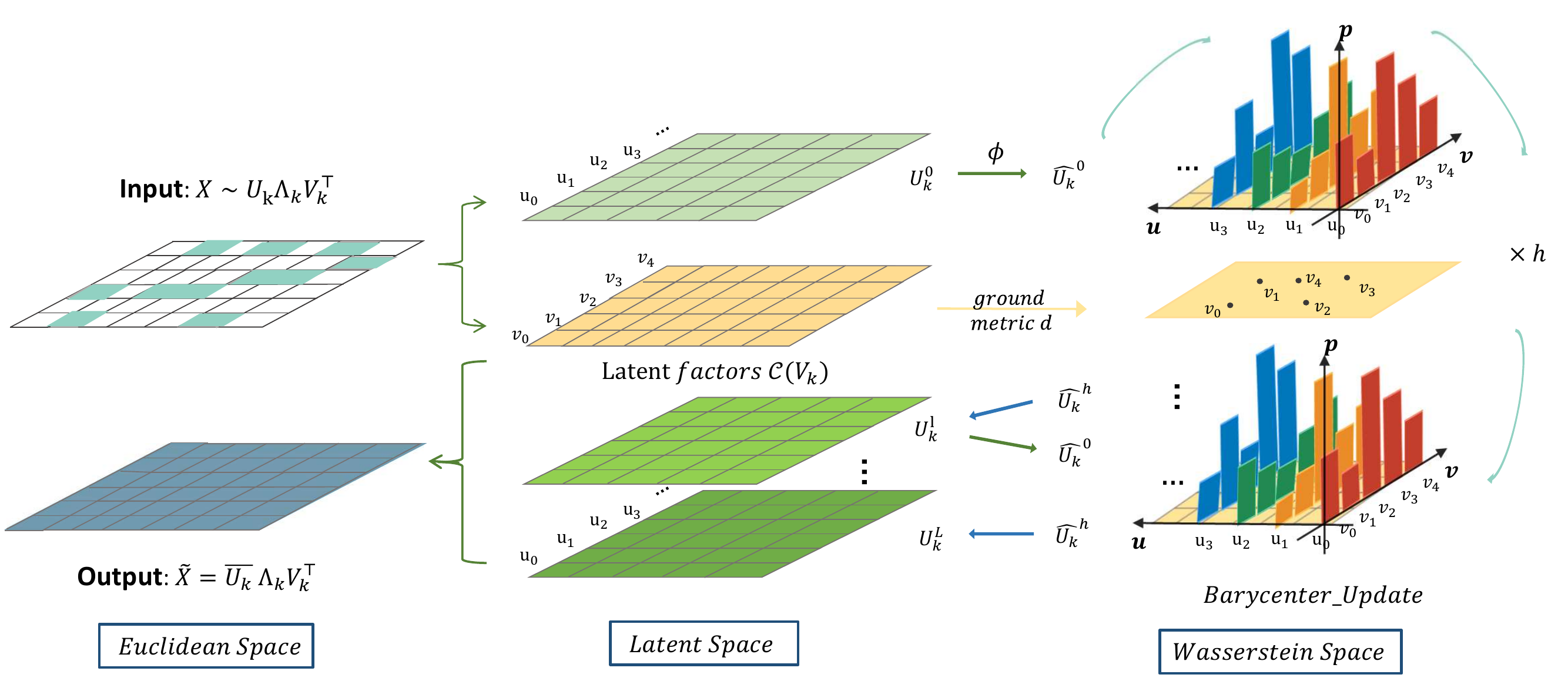}
 \caption{In the \textrm{WGNN} framework, we attempt to incorporate observed node attributes, graph structure and uncertainty caused by missing values in node representations. We first derive the embedding matrix $\mathrm{V}_k$ of $k$ latent factors and weight matrix $\mathrm{U}_k^0$ (middle) from applying low-rank \textrm{SVD} to the incomplete attribute matrix $\mathrm{X}$ (left). With transformation $\phi$, we obtain discrete distributional representations/Wasserstein embeddings $\mathrm{\hat U}_k^0$. To incorporate structural information, we generalize $\textsc{Mean}(\cdot)$ aggregator in Wasserstein space, called \textrm{Barycenter\_Update} for update (right). $\mathrm{\hat U}_k^h$ is the updated Wasserstein embedding. To leverage valid information of $\mathcal{C}(\mathrm{V}_k)$, we pull nodes back to Euclidean space through inverse transformation and use the updated Euclidean embeddings $\mathrm{\tilde X}$ for downstream tasks.}
\label{fig.model}
\end{figure*}

 To handle various graph learning tasks on missing-attribute graphs, we develop an ingenious graph embedding framework, Wasserstein Graph Neural Network (\textrm{WGNN}), which can generate powerful node representations. Although the aforementioned node distributional representations incorporate attribute and uncertainty information, they fail to reflect graph structure information which is essential for graph learning. Inspired by message-passing based GNNs, we adapt the neighborhood aggregation process in GNNs to the node distributional representations. The key idea is to generalize $\textsc{Mean}(\cdot)$ aggregator function to Wasserstein space \cite{frogner2019learning} by computing the Wasserstein Barycenters - the mean of distribution of neighbors for node distributional representation update. In addition, we can pull node distributions back to the original Euclidean space and generate new Euclidean embeddings. Figure \ref{fig.model} depicts the general architecture. \textrm{WGNN} combines the propagation process with a multi-layer perceptron layer (\textrm{MLP}).

To comprehensively investigate the representation ability of \textrm{WGNN}, we design a \textrm{SVD}-based \textrm{WGNN} variant $\textrm{WGNN}_\text{svd}$ for node classification concerning two missing-attribute cases: partially missing and entirely missing. We validate the strengths of our proposed framework on both synthetic and real-world graphs. Extensive empirical results show that compared with all baselines, $\textrm{WGNN}_\text{svd}$ greatly boosts the performance. Furthermore, although our framework is not originally designed for missing values prediction, it can be naturally adapted for the matrix completion task with additional reconstruction constraints. Compared with SOTA matrix completion algorithms \cite{rao2015collaborative,monti2017geometric,hartford2018deep,zhang2019inductive}, our method relies on much fewer parameters and has competitive performance.

\textbf{Contributions.} Overall, our contribution can be summarized as follows: 1. We propose a novel missing-attribute graph learning framework in Wasserstein space, called \textrm{WGNN}, to elegantly generate powerful node representations without explicit data imputation. Our performance is far surpassing that of baselines; 2. We extend \textrm{WGNN} on multi-graph and adapt it for matrix completion with the content of users and items and achieve comparable results of SOTA algorithms with much fewer parameters.

\section{Background and Related Work}

\textbf{Graph representation learning.} In this paper, we focus on learning node representations on attributed graphs. There are many effective graph embedding approaches, such as DeepWalk \cite{bojchevski2017deep}, node2vec \cite{grover2016node2vec}, GenVetor \cite{duarte2019genvectors}, which embed nodes into a lower-dimension Euclidean space and preserve graph structure while most of them disregard node informative attributes. The advent of graph neural networks \cite{bruna2014spectral,kipf2016semi,hamilton2017inductive,velivckovic2017graph,gilmer2017neural,klicpera2019predict} fills the gap, by defining graph convolutional operations in spectral domain or aggregator functions in spatial domain. Although they achieved great success, they highly rely on the completeness and adequacy of attribute information.




\textbf{Machine learning with missing data.} To handle missing data, most machine learning methods rely on data imputation. There is a variety of missing value imputation (\textrm{MVI}) techniques such as mean-filling, \textrm{KNN} imputation \cite{troyanskaya2001missing}, \textrm{softimpute} \cite{hastie2015matrix} with \textrm{SVD}, multivariate imputation \cite{van2007multiple,buuren2010mice}. Also, many deep learning methods are proposed to perform the imputation tasks \cite{gondara2017multiple,yoon2018gain,spinelli2020missing}. One work worth noting is Muzellec {\em et al}~\cite{muzellec2020missing}, which also uses optimal transport to solve the data missing problem. However, it only targets on imputation on general data (without structures) instead of graphs and has a very different methodology. The "imputing before learning" strategy has an important limitation: the performance of models is inherently constrained by the reconstruction ability of the used imputation methods. However, these imputation methods would not always work especially in the extreme missing cases. 

Recently, some advanced models have been developed to directly handle missing data targeting at specified tasks. \textrm{GRAPE} \cite{you2020handling} tackles missing data problems for label prediction and feature imputation. Unlike our work, the missing data is not originally on the graph nodes, but \textrm{GRAPE} represents their two tasks as graph-based problems by leveraging a created bipartite graph. Another recent work, \textrm{SAT} \cite{chen2020learning}, models link prediction and node attribute imputation on missing-attribute graphs with shared-latent space assumption. Different from these works, our \textrm{WGNN} is a graph representation learning framework that focuses on learning node representations with incomplete attribute matrix as input without imputation; and it can be adapted to various downstream graph-based tasks.


\section{Wasserstein Graph Neural Network (\textrm{WGNN})}

In this paper, we propose a graph embedding framework, named Wasserstein Graph Neural Network (\textrm{WGNN}) specified for missing-attribute graphs. \textrm{WGNN} (depicted in Figure \ref{fig.model}) consists of three main components: \textbf{distributional representation generation} to encode observed attribute and uncertainty information, \textbf{Wasserstein aggregation process} to update node distributional representations involving graph structure in a low-dimensional Wasserstein space, and \textbf{Euclidean representation generation} to pull nodes back to a high-dimensional Euclidean space for downstream tasks.

\subsection{Preliminary}
\subsubsection{Notations} 
For a matrix $\mathrm{M}$, we write $\mathrm{M}=(m_i)$, where $m_i$ is the $i$-th row of $\mathrm{M}$. We denote $\mathcal{R}(\mathrm{M})$ as the set of rows of $\mathrm{M}$ and $\mathcal{C}(\mathrm{M})$ as the set of columns of $\mathrm{M}$. Give a missing-attribute graph with a collection of node attributes $\{a_0,\dots,a_{m-1}\}$, we denote observed attribute matrix $\mathrm{X}^{n\times m}=(x_i)$ as its incomplete node attribute matrix, whose missing values are filled with zeros, and $x_i$ is the attribute vector/Euclidean embedding of the $i$-th node.

\subsubsection{Assumption}

We assume that observed values of $\{a_i\}$ come from a collection of $k$ latent factor vectors $\{\nu_0,...,\nu_{k-1}\}$ where $k\ll n$ and each node (e.g. node $j$) may contain attributes from several latent distributions in particular proportions ($\mu_j$). 
Given $\mathrm{X}^{n\times m}$, we assume that there exists low-rank matrices $\mathrm{W}^{n\times k}$ and $\textrm{B}^{m\times k}$ such that $\mathcal C(\textrm{B})$ is the embedding of $k$ latent factors and $\mathcal{R}(\mathrm{W})$ is the weight/probability vectors of nodes affected by latent factors.

\subsubsection{Wasserstein Distance and Wasserstein Barycenter}

\begin{itemize}
    \item \textbf{Wasserstein distance} is an optimal transport metric which measures the distance traveled in transporting the mass in one distribution to match another. The $p$-Wasserstein distance between two distributions $\mu$ and $\nu$ over a metric space $\mathcal{X}$ is defined as 
\begin{align*}
 W_p(\mu,\nu)=\left(\inf_{(x,y)\approx \Pi(\mu,\nu)} \int_{\mathcal{X}\times \mathcal{X}} d(x,y)^p d\pi(x,y)\right)^{1/p}
\end{align*}
where $\Pi(\mu,\nu)$ is the the set of probabilistic couplings $\pi$ on $(\mu,\nu)$, $d(x,y)$ is a ground metric on $\mathcal{X}$. In this paper, we take $p=2$. The Wasserstein space is a metric space that endows probability distributions with the Wasserstein distance. 

\item \textbf{Wasserstein barycenter} of $N$ distributions $\{\mu_i\}$ over $\mathcal{X}$ is an optimizer $\bar \mu$ to the problem:
\begin{align*}
 \bar \mu = \arg\inf_{\mu\in\mathcal{P}(\mathcal{X})}\sum\limits_{i=1}^N W_p^p(\mu,\mu_i),
\end{align*}
where $\mathcal{P}(\mathcal{X})$ denotes the set of all probability measures on $\mathcal{X}$. If $\mu_i$ are discrete distributions supported by $supp(\mu_i)$, support points of barycenter $\bar \mu$ must contain all possible combinations of $supp(\mu_i)$, i.e.:
\begin{equation}
supp(\bar \mu)=\Big\{\frac 1 N\sum_{i=1}^{N}x_i\Big|x_i\in supp(\mu_i) \Big\}.
\end{equation}

\end{itemize}
\subsection{Distributional representation generation}\label{section.space transformation}

 According to the proposed assumption, nodes are mixtures of latent factors and can be expressed as low-dimensional node-factor distributions. In this section, we develop an effective method with matrix decomposition (in our implementation, we use \textrm{SVD}) on the observed attribute matrix to construct a collection of principal latent factors. Thereby, we obtain latent factor embeddings and low-dimensional distributional representations $(\tilde{u}_i)$. 
 The low-rank assumption, which is prevalent in matrix completion, allows missing value imputation methods to obtain a dense and low-rank matrix through dimension reduction techniques. In light of this, we employ \textrm{SVD} to the incomplete attribute matrix $\mathrm{X}$ (similar to \textrm{LSI}\cite{dumais2004latent}). Precisely, through compact \textrm{SVD}, we have $\mathrm{X}=\mathrm{U}\Lambda\mathrm{V}^\top$, where $\mathrm{U}$ is so-called the principal component matrix, $\Lambda$ is a square diagonal matrix with $\Lambda_{ii}=\lambda_i$ ($\lambda_i$ are singular values of $\mathrm{X}$ in descending order) and $\mathrm{V}$ is the basis matrix. Here, we indicate that $\mathcal C(\mathrm{V})$ is a collection of $m$ latent factors. For given $k\ll \min\{n,m\}$, considering the first $k$ principal latent factors, we denote $\mathrm{U}_k=(u_i)$, $\mathrm{V}_k=(v_i)$ and write
\begin{equation}
\mathrm{X}\approx \mathrm{U}_k\Lambda_k \mathrm{V}_k^\top.
\end{equation}
We say $\mathrm{U}_k$ is the weight matrix with respect to $k$ latent factors with $\mathcal C(\mathrm{V}_k)$ as the embedding matrix. 
Allowing negative probability, $u_i$ can be regarded as a generalized discrete distribution supported by $\mathcal{C}(\mathrm{V}_k)$. 
Then we formulate a general transformation function $\phi(\cdot)$ to embed $(u_i)$ to a standard discrete probability space: \begin{equation}\label{wasserstein embedding}
(\tilde u_i) := \big(\phi(u_i)\big) = \big(\varphi(u_i)\big /\parallel \varphi(u_i)\parallel_1\big).
\end{equation}
where $(\tilde u_i)$ is the distributional representation matrix and $\varphi(\cdot)$ is a reversible non-negative function depending on data. In our implementation, $\varphi(\cdot)=\exp(\cdot)$ so that $\phi (\cdot)=\text{sigmoid}(\cdot)$.

\subsection{The Wasserstein aggregation process}
Through the transformation $\phi(\cdot)$, we obtain discrete node-factor distributions $(\tilde u_i)$ which incorporate valid observed attribute information and uncertainty of missing attributes. Noting that Euclidean embeddings fail to express the semantic information of distributions, stemming from the limited expressive capacity of Euclidean space, we take Wasserstein space as the embedding space. Similar to many graph learning methods, we attempt to reflect graph structures in node representations by aggregating information from neighbors.


We first formulate the Wasserstein distance. Denote $supp(\mathrm{U}_k)=\mathcal{C}(\mathrm{V}_k)$ as the set of support points of $(\tilde u_i)$, here we define a ground metric $d$ over $supp(\mathrm{U}_k)$ and obtain the distance matrix $\mathrm{D}^{k\times k}$ as follows:
 \begin{equation} \label{cost metric}
\mathrm{D}_{ij} = d(\mathcal{C}(\mathrm{V}_k)_i,\mathcal{C}(\mathrm{V}_k)_j) =  |\lambda_{i}^2-\lambda_{j}^2|.
\end{equation}
Then we have the corresponding Wasserstein metric:
\begin{equation} \label{wasserstein distance}
W_2^2(\tilde u_i,\tilde u_j|\mathrm{D}) = \min_{\mathrm{T}\geq 0} tr(\mathrm{DT}^\top) \;\;\text{s.t  } \mathrm{T}\mathds{1}=\tilde u_i,\;\mathrm{T}^\top \mathds{1}=\tilde u_j.
\end{equation}

Recall that the $\textsc{Mean}(\cdot)$ aggregator function of neighborhoods in Euclidean space is $\big(\hat u_i^{(l+1)}\big)=\big(\textsc{Mean}\big(\{\hat u_j^{(l)},\forall j \in \mathcal{N}(i)\}\big)\big)$, where $\big(\hat u_i^{(0)}\big)=(\tilde u_i)$ and $\mathcal{N}(i)$ is the neighborhood of node $i$ including $i$ itself. In light of this, we develop $\textrm{Barycenter\_Update}(\cdot)$, a generalized $\textsc{Mean}(\cdot)$ aggregator in Wasserstein space where we update the node distributions by aggregating the neighborhood node distributions. We indicate that Wasserstein barycenter is the mean of distributions $\{\hat u_j^{(l)},\forall j \in \mathcal{N}(i)\}$ in Wasserstein space. Precisely, the formulation of Wasserstein aggregation process in the $l+1$-th \textrm{WGNN} layer is:
\begin{equation}\label{bary_update}
\begin{aligned}
\big(\hat u_i^{(l+1)}\big) &=\text{Barycenter\_Update}\Big(\big(\hat u_i^{(l)}\big),\mathrm{D}\Big)\\
&=\big(\arg \inf_{p\in\mathcal{P}(\mathcal{C}(\mathrm{V}_k))}\sum_{j\in \mathcal{N}(i)}W^2_2(p,\hat u_j^{(l)}|\mathrm{D})\big)
\end{aligned}
\end{equation}
where $\mathcal{P}(\mathcal{C}(\mathrm{V}_k))$ is the set of all discrete distributions supported by $\mathcal{C}(\mathrm{V}_k)$. During the aggregation process, we fix the support of all distributions. That is, we let $supp(\hat u_i^{(l+1)})=\mathcal{C}(\mathrm{V}_k)$ with distance matrix $\mathrm{D}$, for $\forall i,l$. Otherwise, as the aggregation process goes on, $supp(\hat u_i^{(l+1)})$ will be larger and larger and cause high computation complexity (recall that $supp(\hat u_i^{(l+1)})$ should be $\big\{\frac 1 {|\mathcal{N}(i)|}\sum_{j\in\mathcal{N}(i)}x_j\big|x_j\in supp(\hat u_j^{(l)}) \big\}$).  In implementation, we use the Iterative Bregman Projection (\textrm{IBP}) \cite{benamou2015iterative} algorithm to compute such fixed-support Wasserstein barycenter (see Algorithm \ref{Algorithm:IBP}). The complexity of the Wasserstein Barycenter of $N$ $k$-dimensional discrete distributions by \textrm{IBP} is $O(Nk^2/\epsilon^2)$ referring to Kroshnin {\em et al}~\cite{kroshnin2019complexity}. 
The complexity of the Wasserstein aggregation process is $O(|E|k^2/\epsilon^2)$, $|E|$ is the number of edges. In our implementation, the number of iteration $M=100$ and $\epsilon=1e-2$.

\begin{algorithm}[tb]
   \caption{Iterative Bregman Projection}
   \label{Algorithm:IBP}
\begin{algorithmic}
   \STATE {\bfseries Input:} discrete distribution $\mathrm{P}^{d\times n}$, distance matrix $\mathrm{D}^{d\times d'}$, weights vector $w$, $\epsilon$.
   \STATE Initialize $\mathrm{K} = exp(-\mathrm{D}/\epsilon),\;\mathrm{V}_0 = \mathds{1}_{d'\times n}$.
   \FOR{$i=1$ {\bfseries to} $M$}
   \STATE $\mathrm{U}_i = \frac {\mathrm{P}}{\mathrm{KV}_{i-1}}$
   \STATE $\mathrm{V}_i = \frac{exp(log(\mathrm{K}^\top \mathrm{U}_i)w)}{K^\top \mathrm{U}_i}$ 
   \ENDFOR
   \STATE {\bfseries Output:} Barycenter $p_{M}$
\end{algorithmic}
\end{algorithm}

\subsection{Euclidean representation generation}\label{sec.inverse}

Matrix factorization separates the observed attribute information into two parts and stores it in $\mathrm{U}_k$ and $\mathrm{V}_k$. To take full advantage of the information in $\mathrm{V}_k$, we finally pull nodes back to the Euclidean space, which has the same dimension as the original feature space. Thereby, we obtain high-dimensional Euclidean embeddings.
 
We first convert the updated Wasserstein embeddings $(\hat u_i)$ to weight matrix $\mathrm{\bar U}_k$ with $\mathrm{\bar U}_k^\top \mathrm{\bar U}_k=\mathrm{I}_k$ (recall that the initial weight matrix is derived from the orthogonal principal-component matrix), then transform it to the Euclidean embeddings $\mathrm{\tilde{X}}$:
\begin{equation} \label{inver_mapping}
\begin{aligned}
&\mathrm{\bar U}_k = \text{Gram\_Schmidt\_Ortho}\Big(\big(\varphi^{-1}(\hat u_i)\big)\Big),\\
&\mathrm{\tilde{X}} = \mathrm{\bar U}_k \Lambda_k\mathrm{V}_k^\top,
\end{aligned}
\end{equation}
where \textrm{Gram\_Schmidt\_Ortho} is the Gram-Schmidt Orthogonalization processing to maintain the orthogonality of $\mathrm{\bar U}_k$.
In our implementation, $\varphi^{-1}(\cdot)=log(\cdot)$. Interestingly, empirical results show that orthogonalizing node embeddings can efficiently alleviate the over-smoothing problem. Then we can feed $\mathrm{\tilde{X}}$ to arbitrary neural networks to handle various downstream tasks such as node classification. Similar to most \textrm{GNNs}, we equip \textrm{WGNN} with a \textrm{MLP}.

\textbf{Reconstruction for matrix completion.} Obviously, $\mathrm{\tilde{X}}$ is not a matrix completion for $\mathrm{X}$ but a new Euclidean representation matrix. However, with an additional reconstruction constrains, \textrm{WGNN} is able to recover missing values. More details are provided in Section \ref{matrix completion section}.

\section{Empirical Study}

 In this section, we examine \textrm{WGNN} in node classification on synthetic and real-world datasets under two missing-attribute settings to verify the advantages of low-dimensional distributional embedding and Wasserstein aggregation. In addition, we adapt \textrm{WGNN} to matrix completion problems to illustrate the capacity of missing value prediction of our proposed framework.
 
\subsection{Node classification on synthetic data}

\begin{figure}[!ht]
 \centering
\includegraphics[width=0.27\textwidth, angle=270]{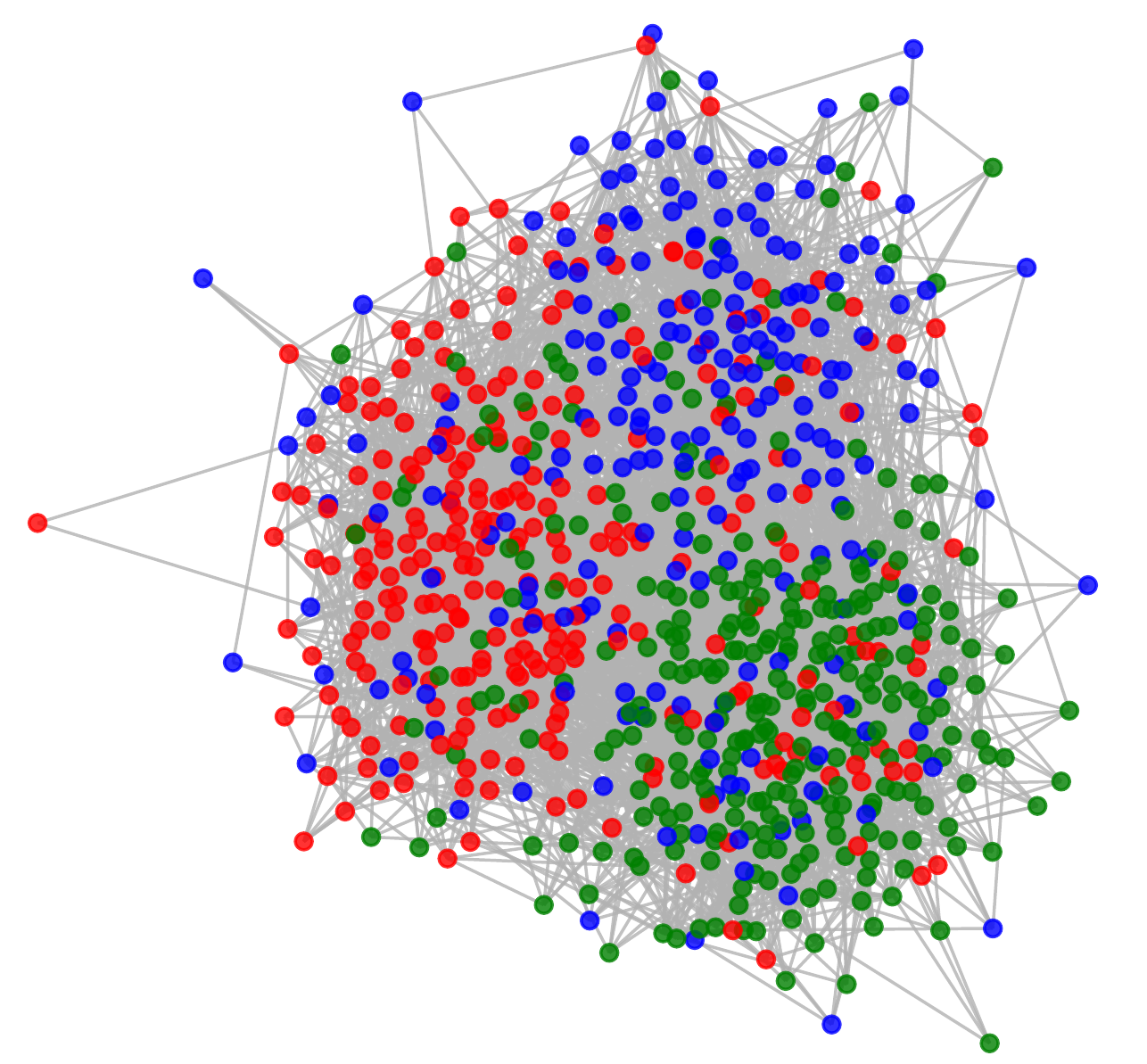}
 \caption{A synthetic partition graph with 3 communities whose sizes are 262, 207 and 319, respectively. 3 labels are assigned to these 788 nodes (colored with red, blue and green) while $85\%$ nodes in a community have the same label.}
\label{fig.synthetic_graph}
\end{figure}

We conduct synthetic experiments in node classification to illustrate the strengths of crucial components of our proposed framework. We first create synthetic data and develop some ablation models as follows.

\subsubsection{Synthetic data}
We generate a random undirected partition graph with three communities whose probabilities of internal and external connections are 0.035 and  0.005, respectively (shown in Figure \ref{fig.synthetic_graph}). We randomly assign nodes with three labels such that $85\%$ nodes in a community have the same label. Nodes with different labels have different families of distributions $\{p_{\theta}(x)\}$ over 100 attributes, i.e., Gumbel distributions, Logistic distributions, and Laplace distributions. Node attribute values are then sampled from their distributions. $\theta$ is the random parameter of these distributions which is different for each node; and $\{p_{\theta}(x)\}$ are exactly the node-attribute distributions that are high-dimensional. Following our proposed assumption, we can also obtain lower-dimensional node-factor distributions by decomposing the attributes through \textrm{SVD}. That is, we have both low-dimensional and high-dimensional node distributional representations of this synthetic graph.

\subsubsection{Missing settings}
Given a full attribute matrix $\mathrm{X}\in \mathbb{R}^{n\times m}$, we generate incomplete input under two missing settings:

\textsc{Partially Missing.} For partially missing rate $r_p\in\{0.1,\dots,0.9\}$, we generate a mask $\mathrm{M}\in \{0,1\}^{n\times m}$ with $p(\mathrm{M}_{ij}=0)=r_p$ and take $\mathrm{X}\odot \mathrm{M}$ as input.

\textsc{Entirely Missing.} For entirely missing rate $r_e\in\{0.1,\dots,0.9\}$, we generate a mask vector $m\in \{0,1\}^{n}$ with $p(m_{i}=0)=r_e$ and take $\mathrm{X}\cdot m$ as input.

\begin{algorithm}[ht]
\caption{$\textrm{WGNN}_\text{svd}$ for node classification}
\label{Algorithm:missing_feature}
\begin{algorithmic}
   \STATE {\bfseries Input:} incomplete matrix $\mathrm{X}$ filled with zeros, $k$. 
   \STATE Apply $k$-rank SVD to $\mathrm{X}$: $\mathrm{X}\approx \mathrm{U}_k\Lambda_k\mathrm{V}_k^\top$; \STATE Initialize $\big(u_i^{(0)}\big) \gets \mathrm{U}_k$  
   \FOR{$l=1$ {\bfseries to} $L$}
   \STATE $\big(\tilde u_i^{(l-1)}\big) = \big(\exp(u_i^{(l-1)})\big /\parallel \exp(u_i^{(l-1)})\parallel_1\big)$
   \STATE $\big(\hat u_i^{(0)}\big) \gets \big(\tilde u_i^{(l-1)}\big)$
   \FOR{$t=1$ {\bfseries to} $h$}
   \STATE $\big(\hat u_i^{(t)}\big) = \big(\arg \inf\limits_{p\in\mathcal{P}(\mathcal{C}(\mathrm{V}_k))}\sum\limits_{j\in \mathcal{N}(i)}W^2_2(p,\hat u_j^{(t-1)}|\mathrm{D})\big)$ 
   \ENDFOR
   \STATE $\big(u_i^{(l)}\big) = \text{Gram\_Schmidt\_Ortho}\Big(\big(\log(\hat u_i^{(h)})\big)\Big)$ 
   \ENDFOR
   \STATE $\mathrm{\tilde X} = \Big(\sum\limits_{l=1}^Lu_i^{(l)}\Big/L\Big) \Lambda_k \mathrm{V}_k^\top$
   \STATE Apply MLP for node classification:$\mathrm{Y} = \text{softmax}\big(\text{MLP}(\mathrm{\tilde X})\big)$ 
\end{algorithmic}
\end{algorithm}

\subsubsection{Testing models} 
\begin{itemize}
    \item $\textrm{WGNN}_\text{svd}$. In node classification tasks, we develop a variant of \textrm{WGNN}, named $\textrm{WGNN}_\text{svd}$, using \textrm{SVD} to generates low-dimensional node distributions (see Algorithm \ref{Algorithm:missing_feature}). We denote $L$ and $h$ are the hyper-parameters of \textit{the number of \textrm{WGNN} layers} and \textit{times of Wasserstein aggregation} in each layer, respectively. We take the average output of each layer as the updated principal-component representations to avoid over-smoothing. 
    
    \item $\textrm{WGNN}_\text{prob}$. In order to verify the superiority of our proposed low-dimensional distributional representation generation method on missing-attribute graphs, we develop an ablation model $\textrm{WGNN}_\text{prob}$ which directly apply the Wasserstein aggregator on the high-dimensional distributional representations $\{p_\theta(x)\}$. 
    
    \item $\textrm{WGNN}_\text{rand}$. To illustrate the effectiveness of latent factors embedded through \textrm{SVD}, we consider a collection of random latent factors. Precisely, we randomly generate a set of $k$ orthogonal vectors $\{b_i\}$ and denote $\mathrm{B}=(b_i)$ as the embedding matrix of random latent factors and obtain mixture weight matrix $\mathrm{W}$ and "singular matrix" $\mathrm{R}$ such that $\mathrm{X}=\mathrm{W}\mathrm{R}\mathrm{B}^\top$. Then we construct another ablation model $\textrm{WGNN}_\text{rand}$ by adapting our space transformation and Wasserstein aggregator to $(\mathrm{B},\mathrm{R},\mathrm{W})$. 
    
    \item $\textrm{GNNs}$ and $\textrm{MLP}$. We also compare $\textrm{WGNN}_\text{svd}$ with some standard baselines: \textrm{MLP} and two \textrm{GNN} models, i.e., \textrm{GCN} \cite{kipf2016variational} and \textrm{GraphSAGE}\cite{hamilton2017inductive}.

\end{itemize}
\subsubsection{Results.} 
\begin{figure*}[!ht]
 \centering
\includegraphics[width=0.65\textwidth]{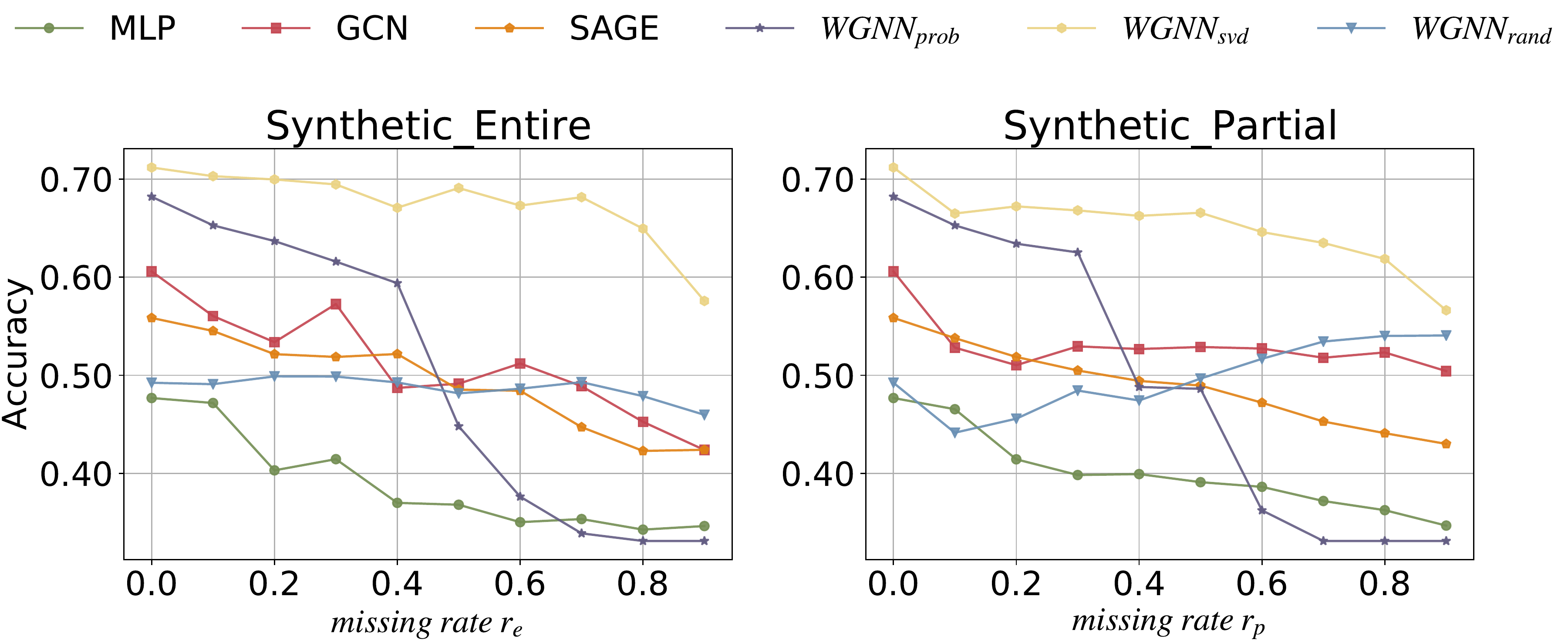}
 \caption{Average accuracy of six testing models node classification on synthetic data under entirely missing (left) and partially missing (right) settings. $\textrm{WGNN}_\text{svd}$ has a huge lead on the synthetic graph with all missing-levels.}
\label{fig.synthethic_result}
\end{figure*}

Experimental results shown in Figure \ref{fig.synthethic_result} demonstrate that: 
\begin{itemize}
    \item When there are no missing attributes, both $\textrm{WGNN}_\text{svd}$ and  $\textrm{WGNN}_\text{prob}$ still outperform other baselines including Euclidean graph embedding methods. It implies the power of Wasserstein embeddings.
    \item At low missing rates, $\textrm{WGNN}_\text{prob}$ has comparable performance with $\textrm{WGNN}_\text{svd}$ while it fails at high missing rates. It shows the superiority of our proposed low-dimensional distributional representation generation method, especially in extremely missing cases.
    \item $\textrm{WGNN}_\text{rand}$ has consistent poor performance across all missing attribute settings which shows the necessity of a proper matrix factorization (e.g. \textrm{SVD}) to figure out appropriate latent factors.
\end{itemize}

\subsection{Node classification on real-world data}\label{ablation study}


\subsubsection{Experimental setup} 

\textbf{Datasets.} We conduct experiments on three Citation datasets: \textsc{Cora}, \textsc{Citeseer} and \textsc{PubMed} (summarized in Table \ref{Citation dataset}).

\begin{table}[t]
\centering
\setlength{\tabcolsep}{0.9mm}{
\caption{Statistics of synthetic data and citation networks.}
\label{Citation dataset}
\begin{tabular}{lccccc}
  \hline
  Dataset & Node &Edge &Feature &Class &Label Rate 
  \\
  \hline
  Synthetic & 788 & 4726 & 100 & 3 & 0.1 \\
  Cora  & 2,708 &5,278  &1,433 & 7 & 0.052\\
  Citeseer  &3,327   & 4,552 & 3,703 & 6 &0.036 \\
  Pubmed  & 19,717  & 44,324 & 500 &3 &0.003 \\
  \hline
\end{tabular}}
\end{table}

\textbf{Baselines.} We first compare against two structure-based methods: the \textrm{Label Propagation Algorithm} (\textrm{LP}) and $\textrm{GCN}_\text{nofeat}$ (take an identity matrix as input). For other graph learning methods, we apply different imputation approaches for data preprocessing, including \textrm{zero-filling}, \textrm{mean-filling}, \textrm{soft-impute} based on low-rank \textrm{SVD} \cite{mazumder2010spectral}, \textrm{KNN-impute} \cite{batista2002study}, and two deep learning techniques: \textrm{GINN} based on a generative adversarial networks \cite{spinelli2020missing}, and \textrm{MIDA} based on a deep denoising autoencoder \cite{gondara2017multiple}. To compare with graph neural networks, we develop six baselines based on \textrm{GCN}: $\textrm{GCN}_\text{zero}$, $\textrm{GCN}_\text{mean}$, $\textrm{GCN}_\text{soft}$, $\textrm{GCN}_\text{knn}$, $\textrm{GCN}_\text{ginn}$, $\textrm{GCN}_\text{mida}$ and four working with \textrm{GAT}: $\textrm{GAT}_\text{zero}$,$\textrm{GAT}_\text{soft}$, $\textrm{GAT}_\text{ginn}$ and $\textrm{GAT}_\text{mida}$. Moreover, following the low-rank assumption, we develop an additional baseline $\textrm{GCN}_\text{svd}$ by feeding \textrm{GCN} with $\mathrm{U}_k\Lambda_k {\mathrm{V}_k}^\top$ instead of $\mathrm{X}$.

It is worth noting that there is a difference between the experimental setting we use for testing \textrm{MIDA} baselines and the original one in our paper. \textit{As \textrm{MIDA} requires training data, we use $\mathbf{30\%}$ nodes whose attributes are complete for training.} That is, with the same missing rate, the number of missing values in the setting of \textrm{MIDA} is only $70\%$ of that in our setting.

\textbf{Model configurations.} For all experiments, we train models using Adam optimizer with 0.01 learning rate. We early stop the model training process with patience 100, select the best performing models based on validation set accuracy, and report the mean accuracy for 10 runs as the final results. We apply two-layer \textrm{MLP} with 128 hidden units for node prediction with fixed $k=64$. The optimal $L$, $h$ and the layers of \textrm{GCN} and \textrm{GAT} differ in different missing level. For partially missing, $L=14$ and $h=2,4,6$ when $r_p\in[0.1,0.3],\;r_p\in[0.4,0.6],\;r_p\in[0.7,0.9]$ respectively. We use 2-layer \textrm{GCN} and \textrm{GAT} when $r_p\in[0.1,0.6]$ and 4-layers when $r_p\in[0.7,0.9]$. For whole missing, when $r_e\in[0.1,0.5]$, $L=10,\;h=2$; when $r_e\in[0.6,0.9]$, $L=14,\;h=6$. We use 2-layer \textrm{GCN} and \textrm{GAT} when $r_e\in[0.1,0.3]$ and 4-layer when $r_e\in[0.4,0.9]$.  

\subsubsection{Experimental results}\label{Sec.Result}
 
\begin{figure*}[ht]
  \centering
    \begin{minipage}[b]{0.97\textwidth}
\includegraphics[width=\linewidth]{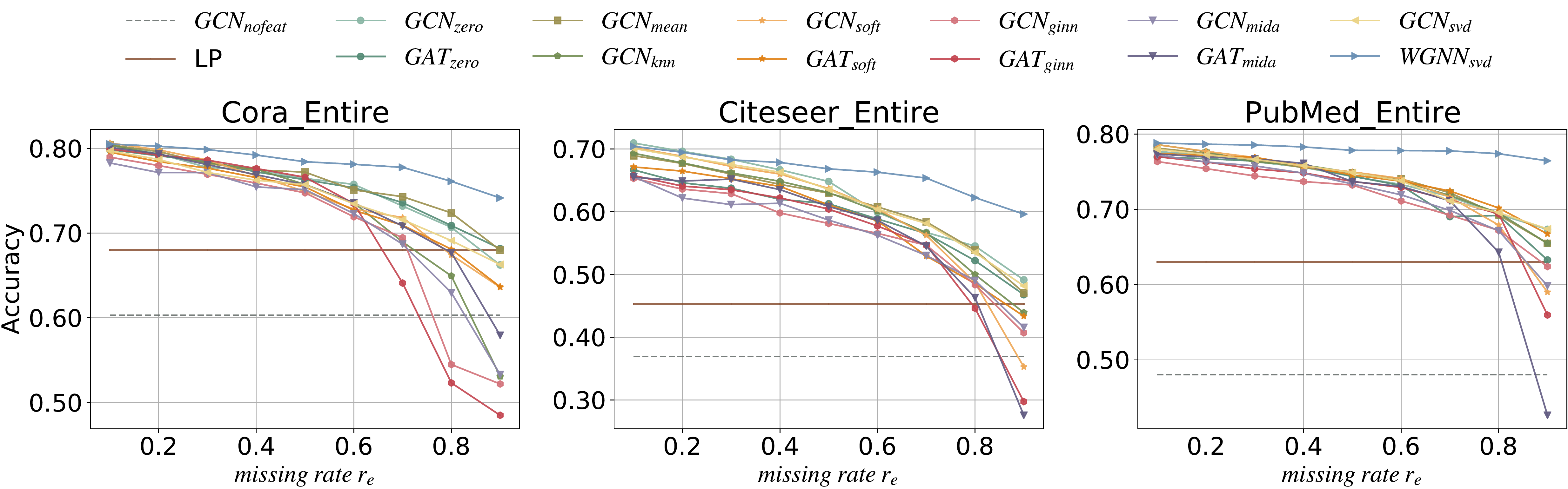}\vspace{4pt}
\includegraphics[width=\linewidth]{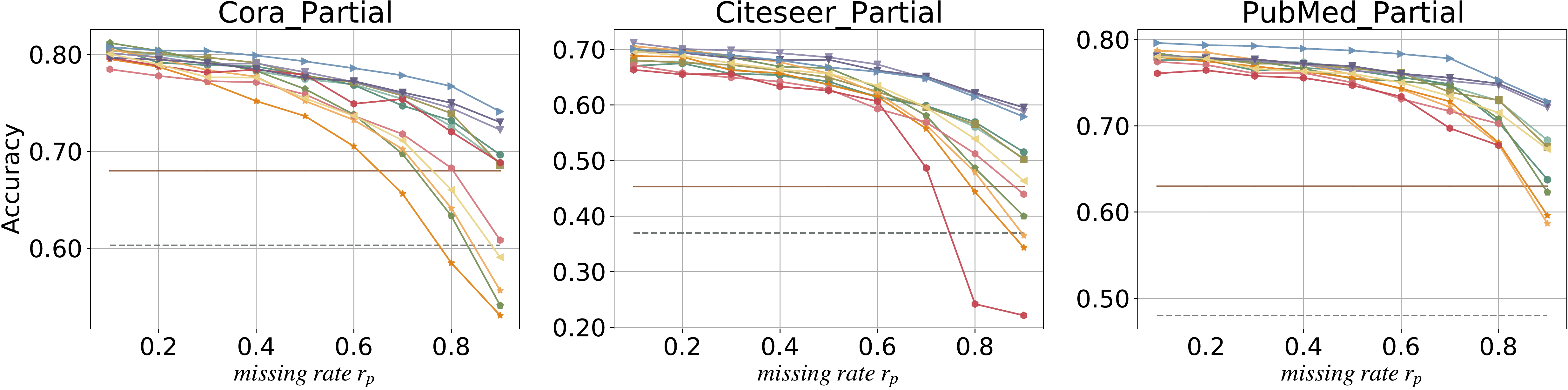}
\end{minipage}
  \caption{Average accuracy of node classification with different \textit{entirely missing} (upper) and \textit{partially missing} (lower) levels. At the highest missing level, $\textrm{WGNN}_\text{svd}$ yields 8\% higher accuracy in entirely missing case and 5\% higher accuracy in partially missing case compared with the best baselines.}
	\label{fig.node_classification}
\end{figure*}
	
As shown in Figure \ref{fig.node_classification}, $\textrm{WGNN}_\text{svd}$ has the best performance on all datasets across all entirely and partially missing levels. Even though \textrm{MIDA} leverages 30\% more observed values, it does not show any superiority, especially in the entirely missing case. Compare with other baselines, $\textrm{WGNN}_\text{svd}$ exhibits significant advantages at high missing level and yields at least 6\%, 10\% and 9\% higher accuracy when $r_e=0.9$ and 5\%, 8\% and 5\% higher accuracy when $r_p=0.9$ over \textsc{Cora}, \textsc{Citeseer} and \textsc{PubMed}, respectively. It implies that \textrm{WGNN} can indeed greatly reduce information distortion by incorporating known semantic information and structure. Conversely, when $r_p=0.9$, several baselines like $\textrm{GCN}_\text{soft}$ and $\textrm{GCN}_\text{knn}$ show lower performance than $\textrm{GCN}_\text{nofeat}$ on \textsc{Cora} and \textsc{Citeseer}. It demonstrates that it is hard to recover missing values and prone to introduce noise when observed information is rare. When observed information is adequate for missing value imputation, such as more than 60\% attributes are available, all methods except for \textrm{LP} and $\textrm{GCN}_\text{nofeat}$, have comparable performance at a low missing level. It indicates a possible limitation of \textrm{WGNN}: for data with rare missing attributes, \textrm{WGNN} is not necessary due to its added complexity. Simple baselines can already generate good results. In general, \textrm{WGNN} shows consistently remarkable performance across all settings.

\subsubsection{Sensitive analysis}
\begin{figure*}[htbp]
\centering
\subfigure[Sensitivity analysis.]{
\begin{minipage}[t]{0.63\textwidth}
\includegraphics[width=\linewidth]{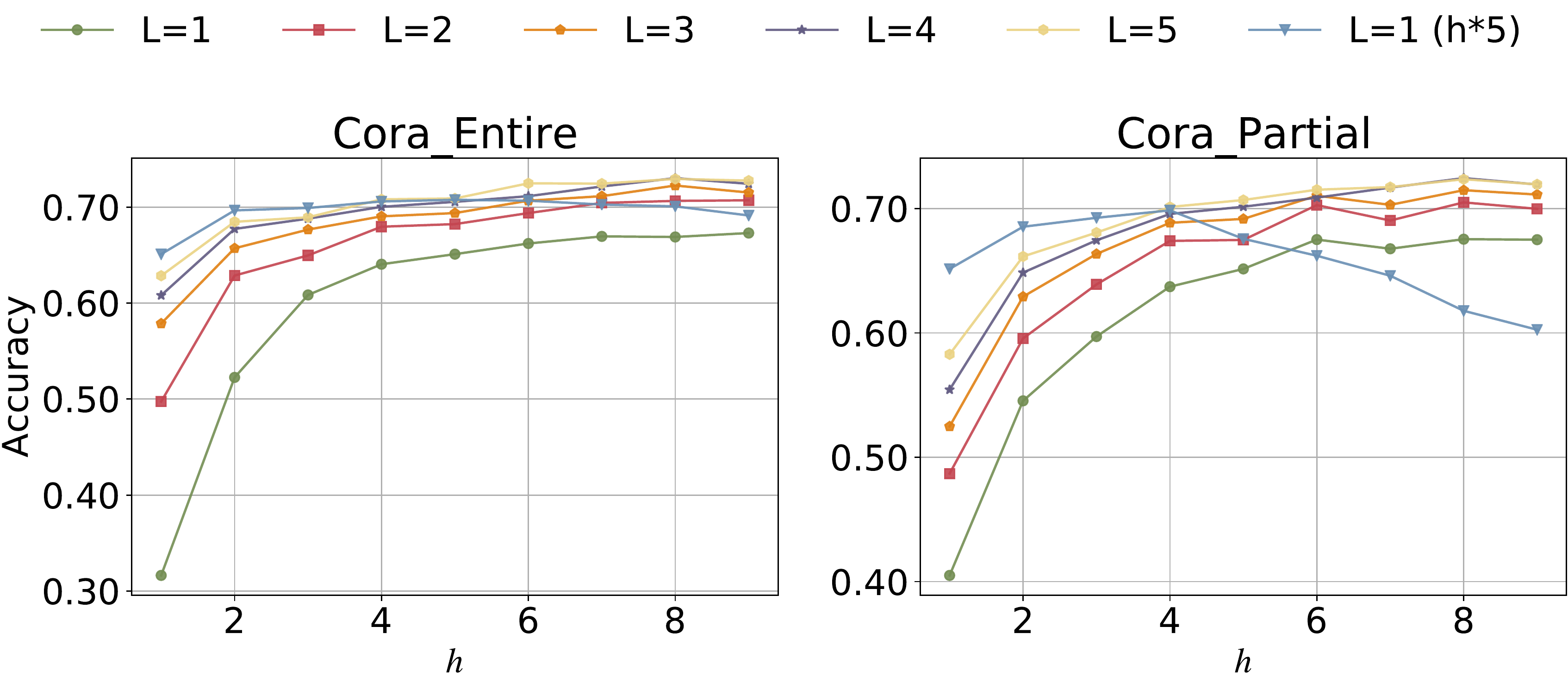}
\label{fig.sensitive analysis}
\end{minipage}%
}%
\subfigure[Ablation study.]{
\begin{minipage}[t]{0.34\textwidth}
\includegraphics[width=\linewidth]{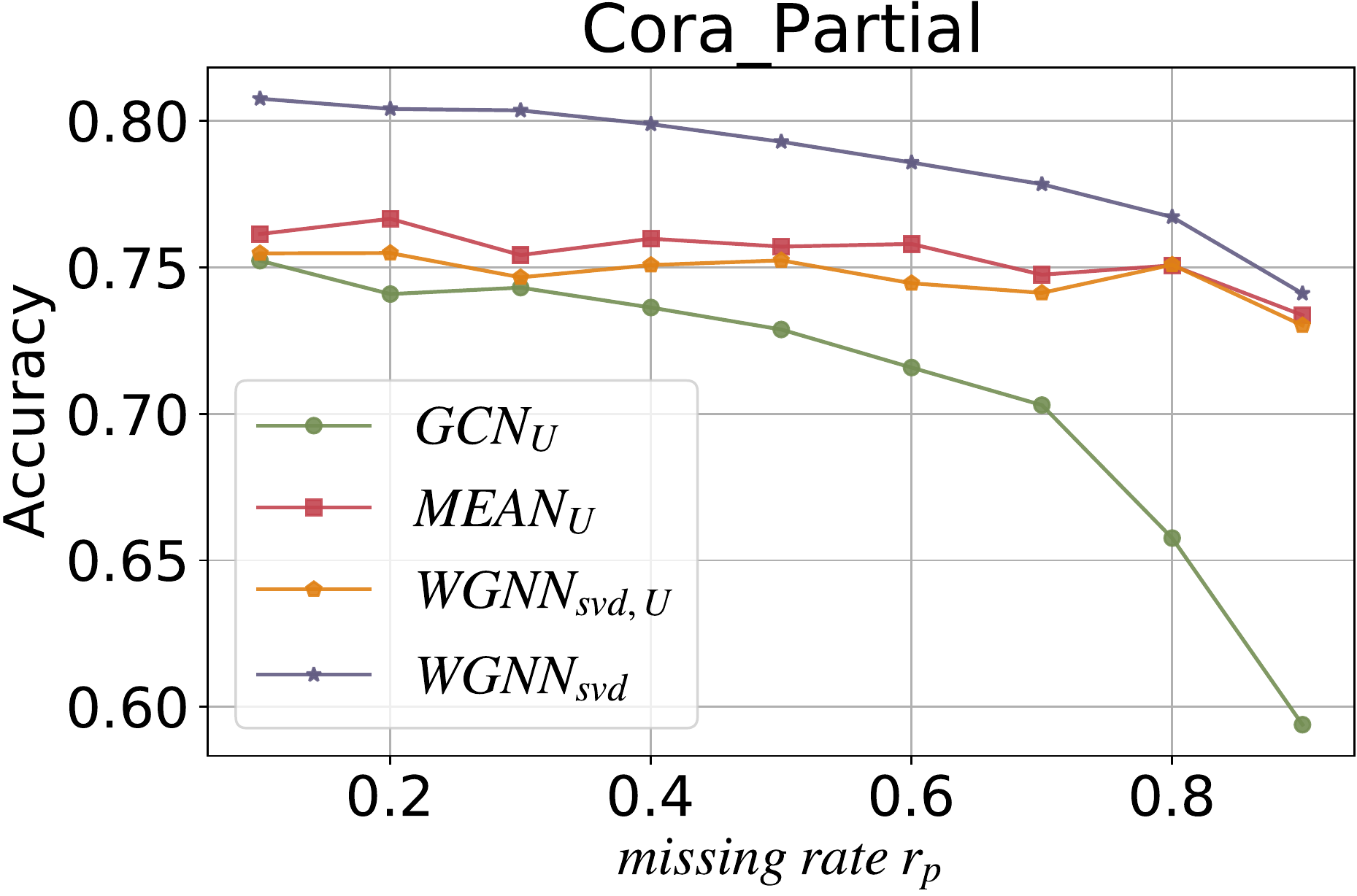}
\label{fig.ablation}
\end{minipage}%
}%
\caption{Sensitivity analysis on times $h$ of Wasserstein aggregation in each layer and the number of \textrm{WGNN} layers $L$ on \textsc{Cora} at $r_e=0.9$ (left) and $r_p=0.9$ (middel), and 
results of ablation models (right).}
\end{figure*}

The results of model tuning are shown in Figure \ref{fig.sensitive analysis}. We only report the experimental results on \textsc{Cora} at 0.9 missing ratio on account of the similar performances on other datasets and missing ratios.

Except for the curve of $L=1(h*5)$, all curves of $L$ have a clear trend of increasing as $h$ increases. Noting that each \textrm{WGNN} layer aggregates information from $h$-hop neighbors, $L$ layers \textrm{WGNN} model involves $L*h$-hop neighbors. $\textrm{WGNN}_\text{svd}$ has the best performance with $L=5, h=9$. In this case, $45$-hop neighbors are involved in the graph embedding process. As we know, many GNN models encounter the over-smoothing issue when they go deep. However, our method can efficiently handle over-smoothing. This is due to two strategies: incorporating the output of all \textrm{WGNN} layers and orthogonalization which make nodes different. The curve of $L=1(h*5)$ provides the experimental illustration: $L=1$ means that there is only once orthogonalization, the performance would decrease starting from $h=4*5$. 

\subsubsection{Ablation study}

To illustrate how \textrm{WGNN} improves the performance, we develop four ablation models to test three main modules of \textrm{WGNN}: \textit{low-dimensional distributional embedding} module, \textit{Wasserstein aggregation} module and \textit{Eucliden embedding} module, respectively. For most ablation models, we report the experimental results on \textsc{Cora} under partially missing settings, shown in Figure \ref{fig.ablation}.

\textbf{\romannumeral1. Low-dimensional distributional embedding module.} In our synthetic experiments, we show the necessity and advantages of \textit{low-dimensional embedding}. Here, we verify it on real-world data. We design an ablation model $\textrm{WGNN}_\text{hd}$ which directly transforms node attributes to discrete distributions $(\hat x_i^{(0)})$ without dimension reduction. Since the semantic information of each dimension of $(\hat x_i^{(0)})$ is unknown, we simply define the ground distance matrix as $\mathrm{\tilde{D}}=\mathbbm{1}-\mathrm{I}$, where $\mathbbm{1}$ is all-ones matrix. Then we generate high-dimensional distributional representations leveraging our Wasserstein aggregator and obtain Euclidean embeddings for label prediction.
 
Even when $r_p=0.1$, $\textrm{WGNN}_\text{hd}$ only has $\mathbf{0.439}$ accuracy on \textsc{Cora}. It is much lower than the worst baseline $\textrm{GCN}_\text{nofeat}$ whose accuracy is 0.6029. It is consistent with our synthetic experiment results. In addition, low-dimensional representations can greatly reduce computation complexity.

\textbf{\romannumeral2.  Wasserstein aggregation module.} By replacing Wasserstein aggregation process with \textrm{GCN} and $\textsc{Mean}(\cdot)$ aggregator to update low-dimensional Euclidean embedding $\mathrm{U}_k$ in \textrm{WGNN} framework, we develop ablation models $\textrm{GCN}_\text{U}$ and $\textrm{MEAN}_\text{U}$ to demonstrate the power of Wasserstein aggregation. 

The architecture of $\textrm{GCN}_\text{U}$ is similar to $\textrm{GCN}_\text{svd}$: 
\[
\mathrm{Y} = \textrm{MLP}\big(\textrm{GCN}(\mathrm{U}_k)\Lambda_k {\mathrm{V}_k}^\top\big).
\]
Here, the updated low-dimensional Euclidean embedding $\textrm{GCN}(\mathrm{U}_k)$ is not orthogonal as Gram-Schmidt Orthogonalization is not differentiable. For $\textrm{MEAN}_\text{U}$, since there is no parameter involved in $\textsc{Mean}(\cdot)$ aggregator, orthogonalization is allowed:
\begin{align*}
&\mathrm{U}_k^{(l)}=\text{Gram\_Schmidt\_Ortho}\big(\textsc{Mean}(\mathrm{U}_k^{(l-1)})\big);\\
&\mathrm{Y} = \textrm{MLP}\Big(\big(\sum_{l=1}^L\mathrm{U}_k^{(l)}/L\big)\Lambda_k {\mathrm{V}_k}^\top\Big).
\end{align*}
As shown in Figure \ref{fig.ablation}, $\textrm{GCN}(\mathrm{U}_k)$ has the worst performance. Interestingly, $\textrm{MEAN}_\text{U}$ has minor fluctuations across all missing ratios. When $r_p=0.9$, its performance is close to that of $\textrm{WGNN}_\text{svd}$. A plausible explanation is that valid semantic information of $\mathcal{C}(\mathrm{V}_k)$ is limited under extreme missing cases. On the other hand, it validates our argument that Wasserstein embedding is more powerful than Euclidean embedding which fails to incorporate semantic information of distributions.
    
\textbf{\romannumeral3. Euclidean embedding module.} \textrm{WGNN} converts the updated representations $\big(u_i^{(l)}\big)$ to high-dimensional Euclidean embeddings for prediction. To demonstrate the advantages of using such Euclidean embeddings, we propose an ablation model $\textrm{WGNN}_\text{svd,U}$ which directly utilize $\big(u_i^{(l)}\big)$ fro prediction:
\[\mathrm{Y}=\text{softmax}\Big(\textrm{MLP}\Big(\big(\sum\limits_{l=1}^Lu_i^{(l)}\big/L\big)\Big)\Big).\]
The curve of $\textrm{WGNN}_\text{svd,U}$ is similar to $\textrm{MEAN}_\text{U}$. It confirms the necessity of incorporating information of $\mathrm{V}_k$ in node representations to improve the accuracy of label prediction.


\subsection{Multi-Graph matrix completion}\label{matrix completion section}
As we mentioned in Section~\ref{sec.inverse}, with additional reconstruction constrains, \textrm{WGNN} enables to reconstruct incomplete matrix. In this section, we test the reconstruction ability of \textrm{WGNN} on recommendation systems with the known pairwise relationship among users and items which is a typical matrix completion problem with user-graph and item-graph.

\begin{algorithm}[ht]
\caption{\textrm{Multi-WGNN} for matrix completion}
\label{Algorithm:matrix completion}
\begin{algorithmic} 
   \STATE {\bfseries Input:} incomplete matrix $\mathrm{X}$ filled with zeros, $k$. 
   \STATE Apply $k$-rank SVD to $\mathrm{X}$: $\mathrm{X}\approx\mathrm{U}_k\Lambda_k\mathrm{V}_k^\top$
   \STATE Initialize $\big(u_i^{(0)}\big) \gets \mathrm{U}_k$, $\big(v_i^{(0)}\big)  \gets \mathrm{V}_k$   
   \FOR{$l=1$ {\bfseries to} $L$}
   \STATE $\big(\tilde u_i^{(l-1)}\big) = \big(\exp(u_i^{(l-1)})\big /\parallel \exp(u_i^{(l-1)})\parallel_1\big)$
   \STATE $\big(\tilde v_i^{(l-1)}\big) = \big(\exp(v_i^{(l-1)})\big /\parallel \exp(v_i^{(l-1)})\parallel_1\big)$
   \STATE $\big(\hat u_i^{(0)}\big) \gets \big(\tilde u_i^{(l-1)}\big)$, $\big(\hat v_i^{(0)}\big) \gets \big(\tilde v_i^{(l-1)}\big)$
   \FOR{$t=1$ {\bfseries to} $h$}
   \STATE $\big(\hat u_i^{(t)}\big) = \big(\arg \inf\limits_{p\in\mathcal{P}(\mathcal{C}(\mathrm{V}_k))}\sum\limits_{j\in \mathcal{N}(i)}W^2_2(p,\hat u_j^{(t-1)}|\mathrm{D})\big)$
   \STATE $\big(\hat v_i^{(t)}\big) = \big(\arg \inf\limits_{p\in\mathcal{P}(\mathcal{C}(\mathrm{U}_k))}\sum\limits_{j\in \mathcal{N}(i)}W^2_2(p,\hat v_j^{(t-1)}|\mathrm{D})\big)$
   \ENDFOR
   \STATE  $\big(u_i^{(l)}\big) = \text{Gram\_Schmidt\_Ortho}\Big(\big(\log(\hat u_i^{(h)})\big)\Big)$
   \STATE $\big(v_i^{(l)}\big) = \text{Gram\_Schmidt\_Ortho}\Big(\big(\log(\hat v_i^{(h)})\big)\Big)$
   \ENDFOR
   \STATE $\big(\hat u_i\big)=\big(\text{Concat}(u^{(0)},\dots,u^{(L)})\big)$
   \STATE $\big(\hat v_i\big)=\big(\text{Concat}(v^{(0)},\dots,v^{(L)})\big)$
   \STATE $\mathrm{\bar U}_k =\big(\text{MLP}_u(\hat u_i)\big/\parallel \text{MLP}_u(\hat u_i)\parallel_2\big)$
   \STATE $\mathrm{\bar V}_k = \big(\text{MLP}_v(\hat v_i)\big/\parallel \text{MLP}_v(\hat v_i)\parallel_2\big)$
   \STATE {\bfseries Return }$\mathrm{\bar X} = \mathrm{\bar U}_k \Lambda_k \mathrm{\bar V}_k^\top$
\end{algorithmic}
\end{algorithm}
 
 \textbf{Multi-WGNN.} For a rating matrix $\mathrm{X}$, $\mathcal{R}(\mathrm{X})$ and $\mathcal{C}(\mathrm{X})$ are attribute vectors of users and items, respectively. Our proposed assumption on the node-graph allows us to express user-nodes as low-dimensional user-factor distributions based on their attributes. Similarly, we can also represent item-nodes as item-factor distributions. Precisely, we generate such distributional representations of users and items, i.e., $\big(\phi(u_i)\big)$ (supported by $\mathcal{C}(\mathrm{V}_k)$) and $\big(\phi(v_i)\big)$ (supported by $\mathcal{C}(\mathrm{U}_k)$), through transformation $\phi(\cdot)$ formulated in Section \ref{section.space transformation}. Then we conduct Wasserstein aggregation in the Wasserstein spaces of users and items for distributional representation update. It is worth noting that, we can perform these processes in parallel without interference since we utilize the fixed-support Wasserstein barycenter for update while the ground metrics of the two Wasserstein spaces defined as Equation \eqref{cost metric} only depend on $\Lambda$. Therefore, this generalized \textrm{WGNN} framework, called \textrm{Multi-WGNN}, can be regarded as an overlay of \textrm{WGNNs}. In the last step we use \textrm{MLP} to optimize the updated $\mathrm{U}_k$ and $\mathrm{V}_k$ then reconstruct $\mathrm{X}$. Algorithm \ref{Algorithm:matrix completion} summarizes the \textrm{WGNN} framework adapted for matrix completion tasks.

\textbf{Benchmarks.} We conduct experiments in the matrix completion task on two popular recommendation systems: \textsc{Flixster} \cite{jamali2010matrix} and \textsc{MovieLens-100K} \cite{miller2003movielens} and use the same preprocessed data and splits provided by Monti {\em et al}~\cite{monti2017geometric}. More Statistics details are provided in Table \ref{completion dataset}.

\begin{table}[tb]
\begin{center}
\caption{Statistics of Flixster and MovieLens-100K.}
\label{completion dataset}
\setlength\tabcolsep{2.3pt}
\begin{tabular}{lllllll}
           Dataset  & Users & Items & Ratings  & Density  & Rating types\\
 \hline   
 \\  
 Flixster  & 3,000 & 3,000   & 26,173 & 0.0029 & 0.5,1,1.5,$\dots$,5    \\    
ML-100K & 943  & 1,682   & 100,000  & 0.0630  & 1,2,$\dots$,5   \\
\end{tabular}
\end{center}
\end{table}

\textbf{Baselines.} We compare our \textrm{Multi-WGNN} model against five advanced matrix completion methods including \textrm{GRALS} \cite{rao2015collaborative}: a graph regularized model utilizing alternating minimization methods and graph structure for completion; \textrm{sRMGCNN} \cite{monti2017geometric}: a geometric matrix completion method applying multi-graph \textrm{CNNs} to graphs of users and items; \textrm{GC-MC} \cite{berg2017graph}: a graph-based method representing matrix completion as link prediction on user-item bipartite graphs; \textrm{F-EAE} \cite{hartford2018deep}: an inductive completion method leveraging exchangable matrix layers; \textrm{IGMC} \cite{zhang2019inductive}: the state of the art matrix completion method using a \textrm{GNN} to enclosing subgraphs for prediction.

\begin{table}[H]
\centering
\caption{RMSE test results on Flixster and MovieLens-100K.}
\label{matrix completion}
\setlength\tabcolsep{2.3pt}
\scalebox{1}{\begin{tabular}{lllllll}
          & GRALS  & sRMG. & GC-MC & F-EAE  & IGMC  & Ours \\
 \hline 
 \\
Flixster  & 1.245 & 0.926  & 0.917 & 0.908 & 0.872 & 0.883    \\
ML-100K & 0.945  & 0.929   & 0.905  & 0.920  & 0.905 & 0.910   \\
\end{tabular}}
\end{table}

\textbf{Experimental settings and results.} We follow the experimental setup of Monti {\em et al}~\cite{monti2017geometric} and take the common metric Root Mean Square Error (\textrm{RMSE}) to evaluate the accuracy of matrix completion. We use 4-layer \textrm{MLP} with 50 hidden units on all datasets. For \textsc{Flixster}, we choose $k=25,\;L=3,\;h=2$. For \textsc{MovieLens-100K}, we set $k=25,\;L=7,\;h=1$. We train the model using Adam optimizer with 0.001 learning rate. We repeat the experiment
10 times and report the average \textrm{RMSE}. The baseline results are taken from Zhang {\em et al}~\cite{zhang2019inductive}. Table \ref{matrix completion} shows the results. As we can see, our \textrm{Multi-WGNN} model has comparable performance as the best baseline \textrm{IGMC}. However, we only rely on a \textrm{MLP}, while \textrm{IGMC} needs to train both \textrm{GCN} and \textrm{MLP}. Furthermore, \textrm{IGMC} requires to extract the enclosing subgraph for each target edge, which is extremely computationally expensive. Therefore, although \textrm{WGNN} is not originally designed for missing value imputation, it shows powerful capacity of reconstruction.

\section{Conclusion}
Missing-attribute graphs are ubiquitous in the real world, while most graph learning approaches have limited ability to leverage incomplete information directly. In this work, we propose \textrm{WGNN}, a framework to generate node representations incorporating observed node attributes, structural information and uncertainty caused by missing values. The key is to transform nodes to a powerful embedding space - a low-dimensional Wasserstein space and develop a suitable message passing schema in such space - the Wasserstein aggregation process. Compared to extensive baselines, our framework shows a significant improvement in prediction, especially when observed information is rare. We further adapt the framework to perform matrix completion with multi-graph. Experiment results on recommendation systems illustrate our capacity to recover missing values.
\bibliographystyle{ACM-Reference-Format}
\bibliography{main}


\end{document}